\documentclass[sigconf]{acmart}
\usepackage{booktabs} 
\usepackage{color}
\usepackage{url}
\usepackage{multirow}
\usepackage{rotating}
\usepackage{array}
\usepackage{color}
\usepackage{wasysym}
\usepackage{verbatim}
\usepackage{bigstrut}
\usepackage{multirow}
\usepackage{amsmath}
\usepackage{setspace}
\usepackage{tabularx}
\usepackage{url}
\usepackage{xspace}
\usepackage{pifont}
\usepackage{subcaption}
\usepackage{array,booktabs}
\usepackage{soul}
\usepackage{vcell}
\usepackage{pifont}
\usepackage{afterpage}
\usepackage{rotating}
\usepackage{wrapfig}
\usepackage{caption}
\usepackage{graphicx}
\usepackage{caption}
\usepackage{enumitem}
\usepackage{xcolor}
\usepackage{balance}
\usepackage[ruled,vlined]{algorithm2e}
\usepackage{algpseudocode}
\usepackage{tikz}
\def\checkmark{\tikz\fill[scale=0.4](0,.35) -- (.25,0) -- (1,.7) -- (.25,.15) -- cycle;} 

\newcolumntype{L}[1]{>{\raggedright\let\newline\\\arraybackslash}p{#1}} 
\newcolumntype{C}[1]{>{\centering\let\newline\\\arraybackslash}p{#1}} 
\newcolumntype{M}[1]{>{\centering\arraybackslash}m{#1}}

\newcommand{\citesec}[1]{Section~\ref{sec:#1}}
\newcommand{\citefig}[1]{Fig.~\ref{fig:#1}}

\newcommand{\citetable}[1]{{Table}~\ref{tab:#1}}

\newcolumntype{R}[1]{>{\raggedleft\let\newline\\\arraybackslash}p{#1}} 
\newcolumntype{L}[1]{>{\raggedright\let\newline\\\arraybackslash}p{#1}} 
\newcolumntype{C}[1]{>{\centering\let\newline\\\arraybackslash}p{#1}} 

\newcommand{\etal}{\emph{et~al.}\xspace}

\newcommand{\mkhmt}{{\it MAPE-K$_{HMT}$}\xspace}
\newcommand{\DR}{{\it Drone Response}\xspace}
\newcommand{\DRx}{{\it Drone Response}}

\definecolor{alizarin}{rgb}{0.82, 0.1, 0.26}

\AtBeginDocument{%
  \providecommand\BibTeX{{%
    \normalfont B\kern-0.5em{\scshape \kern-0.25em b}\kern-0.8em\TeX}}}

\setcopyright{acmcopyright}

\copyrightyear{2022}
\acmYear{2022}
\setcopyright{acmlicensed}\acmConference[SEAMS '22]{17th International Symposium on Software Engineering for Adaptive and Self-Managing Systems}{May 18--23, 2022}{PITTSBURGH, PA, USA}
\acmBooktitle{17th International Symposium on Software Engineering for Adaptive and Self-Managing Systems (SEAMS '22), May 18--23, 2022, PITTSBURGH, PA, USA}
\acmPrice{15.00}
\acmDOI{10.1145/3524844.3528054}
\acmISBN{978-1-4503-9305-8/22/05}

 \usepackage{preprint}
 \markpreprint{SEAMS 2022\\Final published version available at: \url{https://doi.org/10.1145/3524844.3528054}}

 \setcopyright{none}
\renewcommand\footnotetextcopyrightpermission[1]{} 

\begin{document}

\title{Extending MAPE-K to support Human-Machine Teaming}

\author{Jane Cleland-Huang}
\email{janehuang@nd.edu}
\author{Ankit Agrawal} 
\email{aagrawa2@nd.edu} 
\affiliation{%
  \institution{Computer Science and Engineering\\University of Notre Dame}
  \streetaddress{P.O. Box 1212}
  \city{Notre Dame}
  \state{IN}
  \country{USA}
  \postcode{zip}
}

\author{Michael Vierhauser}
\email{michael.vierhauser@jku.at}
\affiliation{%
  \institution{LIT Secure and Correct Systems Lab\\Johannes Kepler University Linz}
  \streetaddress{address}
  \city{Linz}
  \country{Austria}
  \postcode{zip}
}

\author{Michael Murphy} 
\email{murphym18@gmail.com} 
\author{Mike Prieto}
\email{mprieto2@nd.edu}
\affiliation{%
  \institution{Computer Science and Engineering\\University of Notre Dame}
  \streetaddress{P.O. Box 1212}
  \city{Notre Dame}
  \state{IN}
  \country{USA}
  \postcode{zip}
}

\renewcommand{\shortauthors}{Cleland-Huang, Agrawal, Vierhauser, Murphy, Prieto}

\begin{abstract}
 The MAPE-K feedback loop has been established as the primary reference model for self-adaptive and autonomous systems in domains such as autonomous driving, robotics, and Cyber-Physical Systems. At the same time, the Human Machine Teaming (HMT) paradigm is designed to promote partnerships between humans and autonomous machines. It goes far beyond the degree of collaboration expected in human-on-the-loop and human-in-the-loop systems and emphasizes interactions, partnership, and teamwork between humans and machines. However, while MAPE-K enables fully autonomous behavior, it does not explicitly address the interactions between humans and machines as intended by HMT. In this paper, we present the \mkhmt framework which augments the traditional MAPE-K loop with support for HMT. We identify critical human-machine teaming factors and describe the infrastructure needed across the various phases of the MAPE-K loop in order to effectively support HMT. This includes runtime models that are constructed and populated dynamically across monitoring, analysis, planning, and execution phases to support human-machine partnerships. We illustrate \mkhmt using examples from an autonomous multi-UAV emergency response system, and present guidelines for integrating HMT into MAPE-K.
 
\end{abstract}



\maketitle

\newpage\section{Introduction}
The MAPE-K feedback loop~\cite{MAPEK,mapek2}, is a well-adopted reference model for managing and controlling autonomous and self-adaptive systems, and its use has enabled significant advances in autonomous systems over the past decades, for example, in areas such as autonomous driving and traffic management~\cite{gerostathopoulos2019trapped}, Unmanned Aerial Vehicles~\cite{maia2019dragonfly,moreno2019dartsim}, Smart Home and IoT applications~\cite{arcaini2020smart,iftikhar2017deltaiot}, and assistive robots~\cite{jamshidi2019machine}.
Furthermore, rapid advancements in Artificial Intelligence (AI), supported by frameworks such as \mbox{MAPE-K}, have shifted the focus from traditional human-directed robots to fully autonomous ones that do not require explicit human control. These systems, which are commonly developed as ``Human-on-the-Loop'' (HotL)~\cite{fischer2021loop} systems,  differ from ``Human-in-the-Loop'' (HitL) systems in several important ways. In HitL systems, humans make decisions at key points of the system's execution; while HotL systems take full advantage of machine autonomy to perform tasks independently, efficiently, and quickly.

However, given technological advances in autonomic computing, a more advanced form of collaboration, referred to as \emph{Human Machine Teaming} (HMT) has emerged~\cite{HMT}. HMT emphasizes \emph{interactions}, \emph{partnership}, and \emph{teamwork} between humans and machines. It capitalizes upon the respective strengths of both the human and the machine, whilst compensating for each of their potential limitations~\cite{HMT,systems6040044}. According to McDermott~\etal, effective HMT requires \emph{transparency} of the machine's progress and plans, as well as \emph{augmented cognition} to empower the machine to adapt as needed, keep the human partner aware of critical problems, and allow both human and machine to explore the shared solution space. \emph{Coordination} between humans and machines establishes shared knowledge and trust between both human and machine partners and empowers a human partner to direct the machine's behavior when desired~\cite{HMT3,HMT2}. Finally, the machine's self-adaptation capabilities are extended to allow it to dynamically configure and reconfigure its interactions with human partners throughout the mission.

\begin{figure*}[t!]
    \centering
    \includegraphics[width=.95\textwidth]{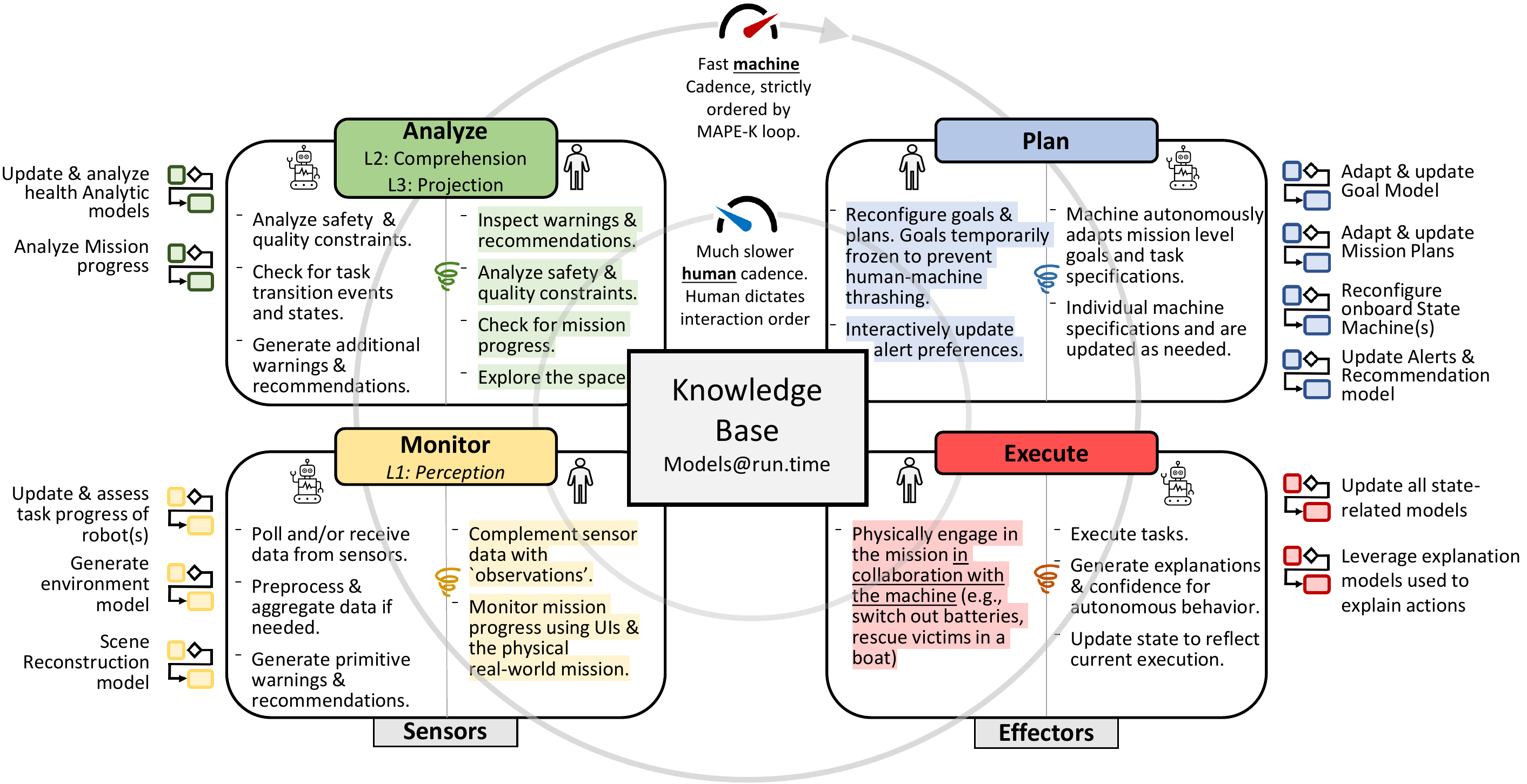}
    \caption{An overview of MAPE-K$_{HMT}$ showing machine activities (outer circle), and human activities (inner circle). Phases are mapped to Situational Awareness levels (L1-L3). Examples of runtime models are shown for each phase.}
    \label{fig:mapek_hmt}
\end{figure*}

HMT systems incorporate aspects of both Cyber-Physical Systems (CPS)~\cite{cps-survey}, and Socio-Technical Systems (STS)~\cite{emery-SST, doi:10.1080/14639220701635470}. In the context of HMT, systems are still expected to operate fully autonomously, with all the capabilities that MAPE-K is designed to support. In fact, not only are the machines capable of performing their tasks autonomously, but they are perceived as true \emph{partners} and not just ``tools'' in achieving mission goals. To make this transition from the HotL paradigm to HMT, humans and machines must interact more closely -- not in a way that reduces or curtails the autonomous behavior of the machine, but in one that leverages that behavior to create meaningful partnerships.

The primary goal of any feedback control system is to remove humans from the loop, and therefore MAPE-K focuses upon autonomous decision-making and self-adaptation without emphasizing the human aspects of a CPS. This was reflected in the results of a recent systematic literature review~\cite{nelly19} which reported that runtime models associated with self-adaptation primarily target the architecture, structure of the system, and/or its goals, but hardly incorporate any human-related factors or activities, such as user interaction or situational awareness~\cite{endsley2012}. This creates a gap between the existing MAPE-K framework and the capabilities needed by an autonomous system to fully interact with human partners in an HTM environment.  Kephart proposed bridging this gap through creating highly interactive ``dialogs'' between humans and machines; however, his examples are all drawn from information systems and not real-time robotic environments~\cite{keynote-Kephart}.

We propose a solution for bridging this gap in more diverse system environments through MAPE-K$_{HMT}$, which enhances the fundamental MAPE-K loop with runtime support for HMT, and aligns \emph{teaming factors}, identified from the HMT literature~\cite{HMT2} with the different phases of the MAPE-K loop. We then present a set of runtime models for use in MAPE-K$_{HMT}$ and describe how they enable support for bidirectional human-machine interactivity and decision-making. 
Our approach is illustrated using a set of worked examples taken from a multi-agent system of autonomous Unmanned Aerial Vehicles (UAV) for supporting emergency response missions~\cite{dronology,DBLP:conf/splc/Cleland-HuangAI20}. Based on these examples, we have derived a lightweight process and recommendations for integrating HMT into MAPE-K.

The remainder of this paper is structured as follows. \citesec{framework} introduces \mkhmt and its extensions to the MAPE-K loop. \citesec{case} introduces our case study system, while~\citesec{application} presents six examples of HMT-related runtime models and their integration into \mkhmt. Section \ref{sec:analysis} then describes a process for utilizing \mkhmt in a MAPE-K system.  Finally, Sections \ref{sec:threats} to \ref{sec:conclusion} present threats to validity, related work, and conclusions.


\section{The \mkhmt Loop}
\label{sec:framework}
With \mkhmt we aim to leverage the benefits of active human engagement while preserving the autonomous behavior of the self-adaptive system. \mkhmt follows the same general structure of MAPE-K, however, as illustrated in \citefig{mapek_hmt}, each of the phases is augmented with additional capabilities targeted to support HMT. The original MAPE-K loop~\cite{MAPEK} consists of four pivotal phases: \emph{Monitoring} in which information is collected from the environment, \emph{Analysis} where data is analyzed to determine if adaptations need to be performed, \emph{Planning} where corresponding actions and adaptations are planned, and finally \emph{Execution} in which the proposed plans are enacted. Additionally, the ``K'' stands for an underlying knowledge base, accessible to all other parts of the MAPE loop, and often supported by runtime models~\cite{uncertainty}.

\begin{table*}[th!]
\centering
\caption{Key impacts of HMT factors upon phases of the MAPE-K loop. The factors (TF1-TF8) are proposed by McDermott~\etal~\cite{HMT3} and augmented to reflect the bidirectional partnerships proposed in this paper. }
\label{tab:hmtfactors}
\addtolength{\tabcolsep}{-1pt}
\begin{tabular}{|C{.42cm}|c| L{1.85cm}|L{4.50cm}|L{8.8cm}|}
\hline
&\multicolumn{2}{l|}{\textbf{Teaming Factor}}   & \textbf{Definition }   &  \textbf{Key example of HMT realization in \mkhmt} \\ \hline



\hline
\multirow{6}{*}{\rotatebox[origin=l]{90}{\bf Transparency}\phantom{xx}}  
&  TF1& Observability                                                                          & Visibility of the task progress of the automated partner and human actions. 
& {\bf [M$^+$]~}Data related to status \& task progress is collected from sensors, UI inputs, and software probes.\newline
{\bf [A$^+$]~}Progress is dynamically visualize for humans. \\\cline{2-5}

& TF2& Predictability   & Transparency of future intentions, states, and activities. 
&{\bf [A$^+$]~}Diverse views generated to provide situational awareness of the machine's activities \& intent.~ {\bf [P$^+$]~}Analysis \& planning results communicated to the human partner.  \\ \hline

\multirow{10}{*}{\rotatebox[origin=l]{90}{\phantom{x}\bf Augmented Cognition}} 
& TF3& Directing Attention & Keeping human partners aware of critical problems through warnings, recommendations, and indicators.
&{\bf [M$^+$]~}Relevant data is collected \& analyzed. \newline{\bf [A$^+$]~}Based on factors such as state or user role, runtime user-alerts  are raised, prioritized, and displayed.
\\ \cline{2-5}

& TF4 &Solution Exploration & Ability for both partners to leverage multiple views, know\-ledge, \& candidate solutions.            
&{ Human partners leverage diverse interactive views and sim\-ulations to explore the solution space {\bf[A$^+$]} and to make plans {\bf[P$^+$]~}}. \\ \cline{2-5}

& TF5&Adaptability                 & Ability to address potentially unexpected evolving, dynamic situations through adaptation. &
{\bf [P$^+$]~}Human engages in adaptation planning alongside machine.\newline
{\bf [E$^+$]~}Machine self-adapts its interactions with humans according to human behavior and context.\\ \hline

\multirow{3}{*}{\rotatebox[origin=l]{90}{\bf Coordination\phantom{xxxxx}} }          
&TF6&Directabilty      & Humans ability to direct/redirect an automated partner's resources, activities, and priorities.  
&{\bf [A$^+$]~}Humans draw on their observations \&  analysis to {\bf [P$^+$]~}intervene in the machine's plans.
{\bf [E$^+$]~}The machine requests support from the human when its  confidence is low. \\ \cline{2-5}

& TF7&Calibrated \newline Trust            & Trustworthiness indicators of machine's ability to make correct decisions in current context.   
&{\bf [M$^+$]~}Based on collected data {\bf [M$^+$]~}the machine {\bf [P$^+$]~} computes reliability of its own  autonomous decisions and actions, and displays appropriate trust-related indicators at runtime {\bf [E$^+$]}.               \\ \cline{2-5}

& TF8&Common\newline Ground             & Shared beliefs, assumptions, and intents across human and automated partners.    
&{\bf [M$^+$]}~The Human understands data collection,analysis \& use, {\bf [A$^+$]}, \&  the machine's capabilities {\bf [P$^+$/E$^+$]}.~~\newline
{\bf [E$^+$]}~The machine responds to human directions as expected.\\ \hline 
\end{tabular}
\end{table*}

 In the context of human-machine teaming, a number of capabilities are required to ensure both successful human-machine interaction and cooperation~\cite{HMT2}. In Table \ref{tab:hmtfactors}, we summarize some of the primary dependencies. 
 \emph{Transparency} is supported by ``Observability'' (TF1) of the autonomous partner's task progress and ``Predictability'' (TF2) of its future plans. \emph{Cognition} is augmented through ``Directing Attention'' (TF3) to critical problems, for example by raising meaningful alerts to increase situational awareness. ``Solution Exploration'' (TF4) and ``Adaptability'' (TF5) imbue both the machine and humans with the knowledge and capabilities they need to make and enact decisions. Finally, \emph{Coordination} is supported through ``Directability'' (TF6), ``Calibrated Trust'' (TF7), and establishing ``Common Ground'' (TF8) to enable  informed, trustworthy, and trusted partnerships. Achieving each of these capabilities requires  consideration for how humans and machines can work in teams to accomplish their goals, then establishing runtime models to collect, aggregate, and visualize information that supports HMT. As a result, human and machine collaborators can engage in meaningful interactions. 
 These eight capabilities represent Teaming Factors that are applied across the  MAPE-K loop to augment each phase as M$^+$, A$^+$, P$^+$, E$^+$, and K$^+$ respectively.  In the following sections, we explain how each phase is augmented to support HMT.
 
\subsection{The Monitoring Phase (M)}
 In the MAPE-K loop, monitoring is primarily concerned with \emph{collecting data} from the self-adaptive system and the environment in which it operates. Hardware sensors provide raw data such as temperature, distance to potential obstacles, video streams, or GPS locations, while software probes provide data from the running system such as its  resource usage, response times, and currently executing tasks~\cite{iglesia2015mape}. This data is collected, persisted, and used in runtime models to guide subsequent analysis and self-adaptation decisions~\cite{vierhauser2018monitoring,jiang2018data}. 
 
 \noindent\textbf{M$^{+}$:} To forge effective human-machine partnerships, HMT environments must provide bidirectional situational awareness; therefore, the machine not only collects data about its own state but also collects human-initiated data reflecting human goals, directives, workload, and response times~\cite{endsley2012}. This data is collected via Graphical UIs (GUIs) and via hardware interfaces such as radio controllers (RC), audio devices, pointing devices, and even eye-trackers \cite{fridman2018cognitive, palinko2010estimating,human-monitoring,8787082} or brain interfaces \cite{gaume2019cognitive,nourmohammadi2018survey}. Furthermore, given the dissonance between the way humans and machines perceive the world~\cite{dissonance2019}, supplementary data is needed to support human awareness and interactions and must be collected from all three sources (i.e., machine, environment, and human inputs). The data collection process is therefore expanded accordingly and used across subsequent analysis, planning, and execution phases  to support the HMT goals of transparency, cognition, and coordination, with an emphasis on the teaming factors of \emph{observability}, \emph{directing attention},  \emph{calibrated trust}, and \emph{common ground}.

\subsection{The Analysis Phase (A)} The MAPE-K analysis phase is concerned with determining whether adaptation actions are required, based on the current and predicted state of the system, the environment it is operating in, and its defined goals, safety constraints, and quality of service specifications. Automated analysis enables timely and fast reactions to changes in the environment and emergent situations.

\noindent\textbf{A$^{+}$:} HMT recognizes the value of augmenting machine analysis capabilities with human perspectives and inputs~\cite{dissonance2019}. This requires information to be exchanged between machines and humans through human-facing interfaces that provide insights into the autonomous machine's intent, performance, plans, and reasoning processes'~\cite{chen-transparency,sanneman2020}. Human analysis often relies upon a combination of \emph{real world} observations as well as \emph{software supported} interactions. In direct real-world analysis, for example, the human uses their physical senses to observe and analyze the machine's behavior, and to intervene quickly and directly through either a hardware or software interface, potentially causing a temporary interrupt to the MAPE-K loop. In contrast, in software-supported analysis, the human analyzes a representation of the machine's behavior and/or state through a GUI. As humans need time to understand and analyze the information and formulate decisions, static snapshots become stale within seconds -- even milliseconds. This leads to the conclusion that GUI support for HMT must include dynamic runtime views that reflect current system states and historical information about past actions, rather than just static snapshots of the system. They are therefore often built on top of existing runtime models with continually updated views. Historical views that enable the human to explore \emph{why} a machine made a past decision (e.g., \cite{Sakizloglou2021IncrementalEO,DBLP:conf/models/ReynoldsGB20}), or \emph{what} it plans to do next, are also needed. The \mkhmt analysis has a broad impact across almost all of the teaming factors as analysis is a precursor to human engagement.

\subsection{The Planning Phase (P)}
\label{sec:planning} In the planning phase the machine plans self-adaptation actions such as switching states to perform different tasks, reconfiguring existing features, activating or deactivating sensors, or modifying polling frequencies to preserve power or to collect additional information about the system or its environment. 

\noindent\textbf{P$^{+}$:} HMT introduces two additional considerations to the planning phase. First, the human leverages their observations of the machine, its operating environment, interactions with other team members, and profound experience as a ``human knowledge base'' to engage directly in the planning process. Humans might reconfigure the mission's goals or plans, or temporarily intervene in the operation of the machine, for example by assuming manual control of a task that the machine is not able to perform autonomously or when the machine malfunctions. 
However, this introduces a potential tug-of-war that can occur when humans and machines create competing plans \cite{delemos}. This was catastrophically illustrated in the crash of Lion Air Flight 610 and Ethiopian Airlines Flight 302 in which the MCAS (Maneuvering Characteristics Augmentation System) incorrectly perceived the angle of attack to exceed predefined limits and therefore pushed the nose of the plane down, whilst pilots struggled to push it back up~\cite{boeingeport}. The system was not designed to detect and mitigate this type of tug-of-war, and ultimately the machine ``won'',  causing the planes to crash. Achieving effective coordination between humans and machines is a challenging problem which we discuss further in  \citesec{coordination}.

HMT has a second major impact on the planning phase, as the machine may make additional self-adaptation plans targeted at enhancing the human's interactive experience. For example, the machine might self-adapt its internal alert system to adjust the type and frequency of human alerts if the system perceives that human response time is starting to lag \cite{ankit2021}. 

\subsection{The Execution Phase (E)} During the execution phase, the previously generated plan or adaptation strategy is executed on the physical machine or device.

\noindent\textbf{E$^{+}$:} In an HMT setting, both the machine and the human partners enact plans -- sometimes closely coordinating their work whilst at other times working more independently on tasks that each partner is best suited to perform. 

\subsection{The Knowledge Base (K) + Runtime Models} 
MAPE-K employs diverse runtime models that represent the  structure, behavior, and/or goals of a system at runtime. These models are pivotal for guiding autonomous adaptation decisions~\cite{Assmann2014}. In a recent systematic literature review Bencomo~\etal, \cite{nelly19} reported that {\it Models@run.time} have been used in numerous ways -- for example to depict the current state of the system \cite{EsfahaniYCM16} and its behavioral dynamics specifying exactly what the system is able to do from its current state \cite{filieri}, model system goals \cite{vialon,castaneda, sabatucci} and functional and non-functional requirements \cite{privacy, qualities}, and to depict product variability \cite{Weibach2017DecentrallyCE}. They provide a bidirectional reflection layer, such that changes in the runtime model trigger changes in the goals, structure, and/or behavior of the underlying system, whilst changes in the system are reflected in the model. As such, runtime models support system autonomy, imbuing the system with the ability to sense, analyze, predict, and to make independent decisions. 

\noindent\textbf{K$^{+}$:} In \mkhmt runtime models must also support the HMT goals of transparency, augmented cognition, and coordination between human and machine partners. Depending on the type of system, application domain, or the type of missions that are executed, this requires different types of runtime models that are explicitly designed to provide information to the user or relay critical information from the user to the system so that informed adaptation decisions can be made.

\section{Case Project: A Multi-UAV System}
\label{sec:case}
Throughout this paper, we illustrate MAPE-K$_{HMT}$ with examples drawn from our {\it \DR} system -- a flexible and configurable framework for multi-UAV missions~\cite{dronology,DBLP:conf/splc/Cleland-HuangAI20}. 
\subsection{The Drone Response Ecosystem} \DR is fully deployable with both physical and simulated UAVs ~\cite{jmavsim,Koenig200}. 
The \DR architecture includes diverse user interfaces, a  Ground Control Station (GCS), and autonomy capabilities located onboard each UAV. The MAPE-K infrastructure is  distributed across the \DR ecosystem as follows. 
 \vspace{2pt}

 \noindent{$\bullet$~}{\it UAV Onboard Pilot:~} The \textit{Onboard Pilot} module acts as an application layer for the autopilot stack which in turn includes flight control software and hardware for executing plans. Its internal \textit{State Machine} receives mission specifications and instantiates itself accordingly. It then uses its onboard state machine to progress through a series of tasks, with transitions triggered by events it detects by analyzing its own sensor data and/or messages received from the GCS or other UAVs. MAPE-K's monitoring, analysis, planning, and execution occur onboard each UAV with an emphasis on managing the tasks and adaptations of each individual UAV. \vspace{2pt}
 
\noindent{$\bullet$~}{\it Ground Control Station:~} The Control Station adheres to the microservices and publish-subscribe architectural style~\cite{muccini2018iot}, with each microservice providing a specific capability such as airspace leasing, multi-UAV coordination, or safety checking mission specifications. During a mission, microservices receive messages from both the human operators and UAVs, construct runtime models, and leverage them for analysis and planning purposes. The knowledge base is thereby distributed across runtime models managed by various microservices and supported by a shared in-memory database. Status data (e.g., GPS location, battery, health) and task progress updates (e.g., current task, potential adaptations) sent to the GCS by both humans and UAVs support MAPE-K's mission-level monitoring, analysis, and planning. \vspace{2pt} 

\noindent{$\bullet$~}{\it Graphical and Hardware Interfaces:~}
\DR uses graphical and hardware user interfaces to enable human-machine interactions. Most of the interactions use GUIs; however, in case of emergency, or to temporarily assume control for tasks that the machine has not yet been trained to perform, humans can directly issue commands to UAVs via hand-held radio controllers. The GUI components are built on a centrally hosted {\it web application} and asynchronously send and receive status data and video streams over a mesh-radio via GCS's message broker and the onboard pilot modules~\cite{ankit2021}. Many of the \mkhmt runtime models are coupled with one or more GUIs that provide situational awareness to the human for planning and analysis purposes while also potentially monitoring aspects of human interaction behavior.\vspace{-6pt}\\ 

Within \DR, HMT is supported by diverse runtime models integrated across the system (cf. \citefig{model}). 
For example, in order to generate meaningful explanations where multiple UAVs are involved, information about their current state is collected from each of the UAVs. Humans interact directly with the UAVs using physical devices (1) and indirectly via a GUI (2). HMT Interaction Models (4) that process data and control human-machine interactions in the GUI (3), receive input from underlying runtime-models (5), as well as from other sources such as UAVs (7) and humans via the GUIs. Communication between these major components is achieved using the MQTT message broker with service level agreements guaranteeing fast response times when needed (6). 

  \begin{figure}[t!]
     \centering
     \includegraphics[width=1\columnwidth]{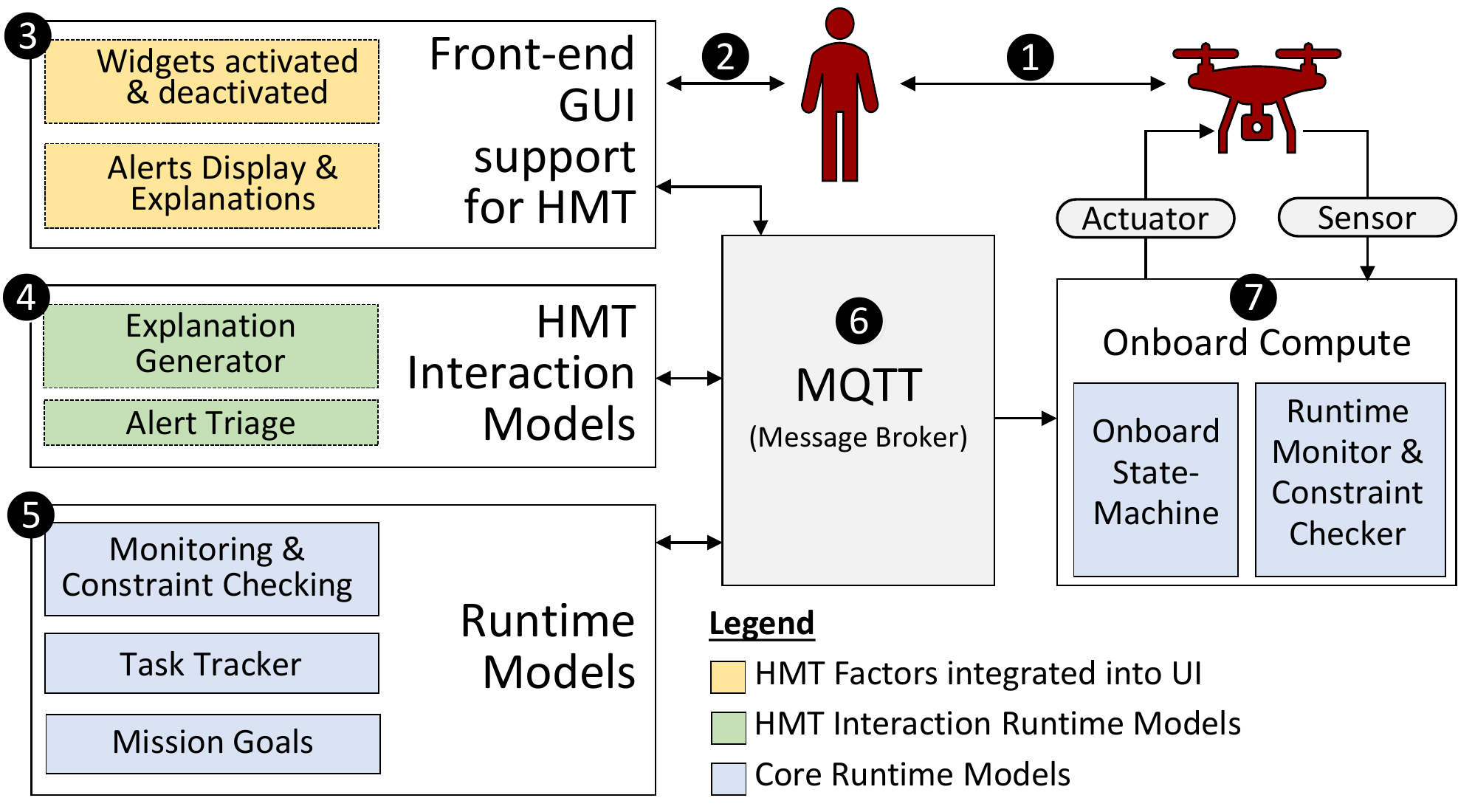}
     \caption{Human-Machine Teaming is supported by GUIs, hardware UIs, and a set of closely integrated runtime models.}
     \label{fig:model}
    \end{figure}

\subsection{A Motivating Scenario}
\begin{figure}[b]
     \centering
     \includegraphics[width=1\columnwidth]{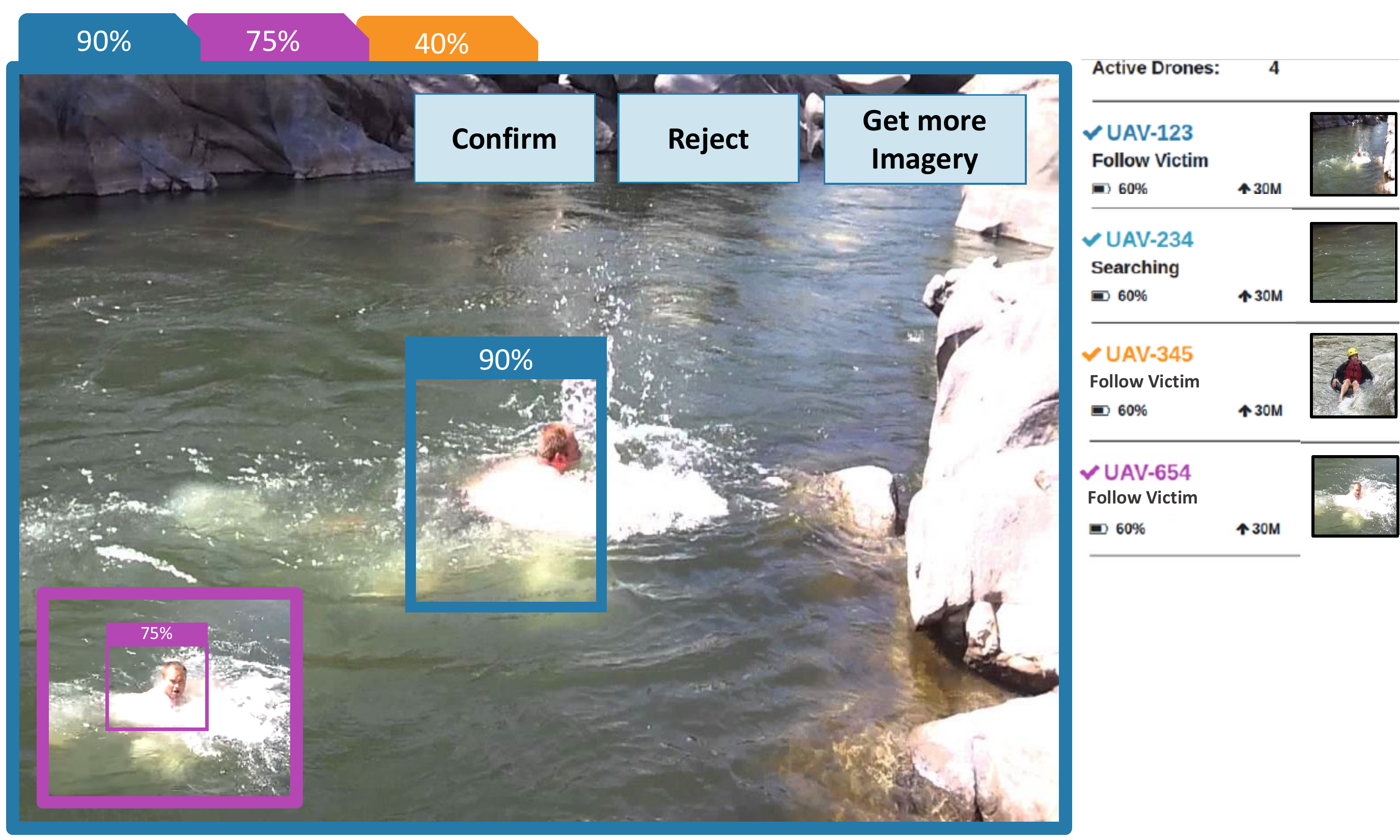}
     \caption{\DR streams video of victims once detected by the onboard Computer Vision.}
     \label{fig:rescue}
     \vspace{6pt}
     \includegraphics[width=1\columnwidth]{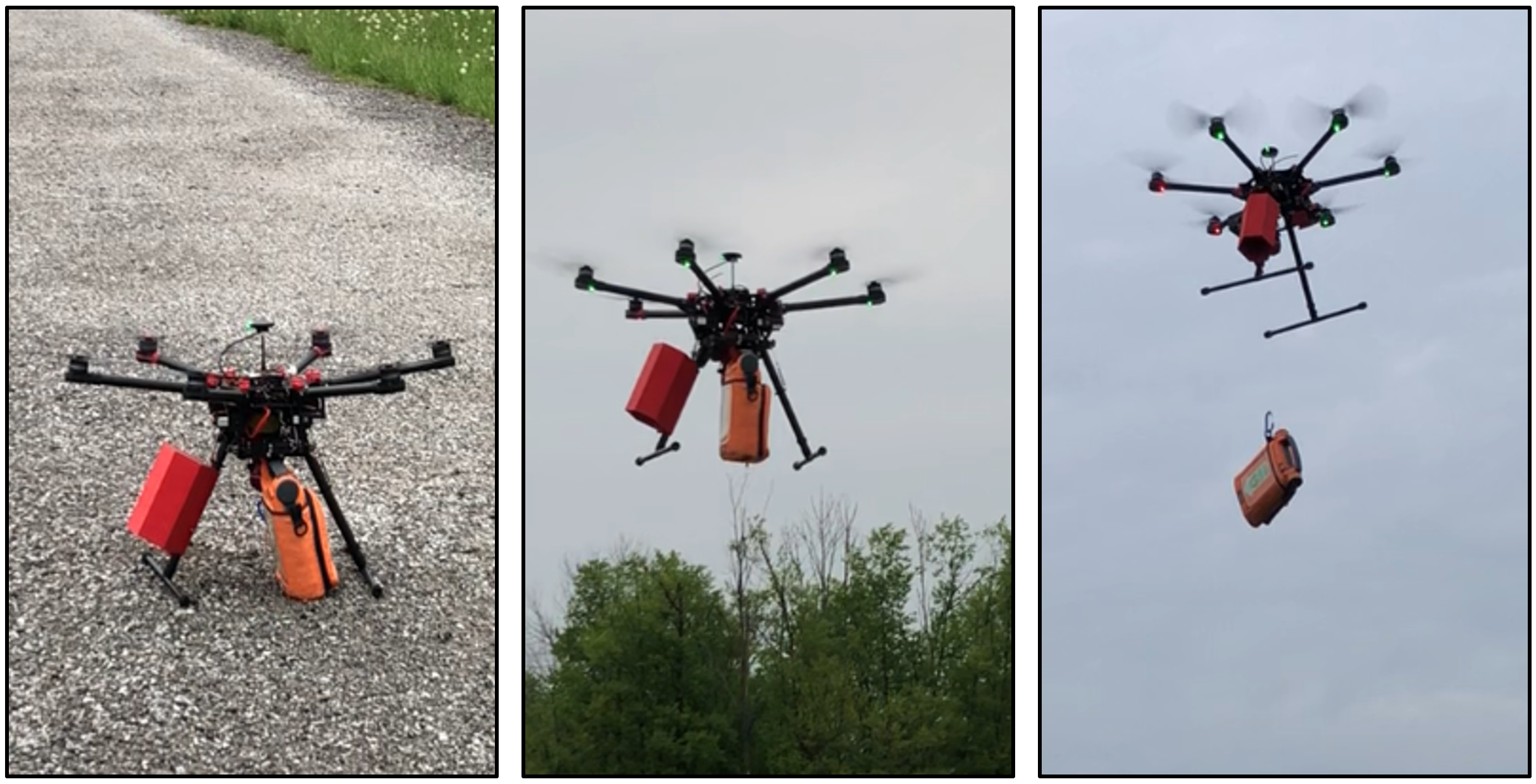}
     \caption{Proof-of-concept \DR delivery of a defibrillator. (See \url{https://youtu.be/AleVXc3QWIk}) }
     \label{fig:delivery}
     
\end{figure}

\DRx's UAVs are designed to leverage computer vision and work alongside humans in emergency response missions such as search-and-detect, surveillance, and rescue scenarios. As depicted in \citefig{rescue}, the UAVs first use their onboard computer vision to search for, and potentially detect the victim. They then notify the human responders about the victim's location. \DR then has the option of delivering a flotation device, as shown in \citefig{delivery};  however, this requires very close human-machine teaming. For example, if a physical rescue from a boat is imminent, then simultaneously dropping a flotation device in the location of the victim could hinder the rescue operation and introduce a potential safety concern. This example provides a clear differentiation between HiTL, HoTL, and HMT paradigms. In a HiTL environment, the UAV would wait to be dispatched by a human and subsequently request input regarding specific tasks to be performed. In a HoTL environment, the UAV might decide to initiate the delivery of the flotation device independently; however, the human could decide to intervene and cancel the  operation. In contrast, while human intervention is definitely part of HMT, the HMT scenario calls for joint decision-making, whereby the human and machine leverage a shared set of beliefs and understanding of the mission to evaluate their capabilities and opportunities, weigh  the options with respect to global mission goals, and ultimately make a joint decision for the good of the mission.
 
All eight HMT teaming factors (cf. Table \ref{tab:hmtfactors}) play a vital role in fostering collaborations between emergency responders and the UAVs. Because humans need to develop trust in the autonomy capabilities of the UAV during the mission, they require observability of their tasks, and expect predictability  of their decisions and actions. The system must raise appropriate alerts in order to help humans to maintain situational awareness. 
In the spirit of the HMT partnership, the human can directly contribute  to the search  -- for example, providing information for use by the computer vision algorithm or hints to the route planner, that the victim is wearing a red sweater and that sightings have been reported in a certain part of the search area.  To increase the likelihood of a successful search, the UAVs and humans must establish a shared conceptual model of the mission  and humans must be able to explore the solution space  through user-facing perspectives of the current mission, including the location, status, and progress of each UAV.

\section{Applying MAPE-K$_{HMT}$}
\label{sec:application}

\begin{figure}[t]
     \centering
     \includegraphics[width=.98\columnwidth]{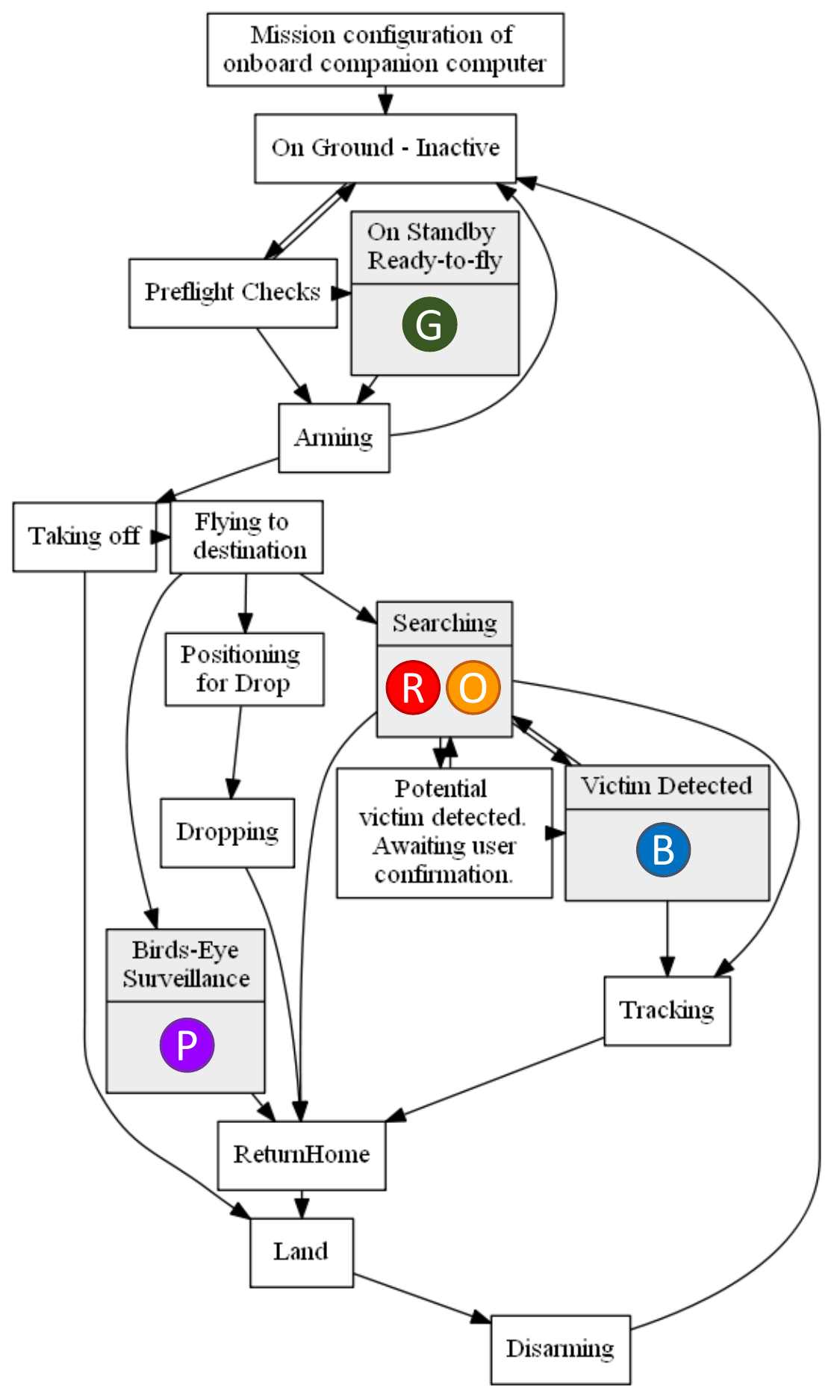}
     \caption{This `Multi-Agent Tracking' view displays the task progress of all active UAVs. It is managed by a dedicated microservice, which continually aggregates states and transition paths for all active UAVs and uses each UAV's uniquely colored token to mark their current state.}
     \label{fig:state_diagram}
     \vspace{-12pt}
\end{figure}
In this section, we describe six different runtime models from our \DR project that support HTM within the MAPE-K loop.
Each of the runtime models contributes towards one or more HMT goals of transparency, augmented cognition, or  human-machine coordination. Furthermore, each model not only reflects the runtime behavior or structure of the machine, but also provides direct or indirect support for a human-facing UI component in order to support human-machine collaboration.  In the following sections we describe models that are particularly pertinent to transparency, cognition, and coordination respectively. 

\subsection{Runtime Models for Transparency}
HMT's transparency goal focuses on observability and predictability and therefore aligns closely with Endsley's Situational Awareness goals of {\it observing} and {\it understanding} ~\cite{endsley2012,endsley2017autonomous}. Observability means that humans are aware of what their autonomous partner is doing, including its goals, current tasks, future intentions, status, progress, ability to adapt to changing contexts, rationales for adaptations, and challenges or constraints that impact its ability to solve the current problem.
Additionally, {\it predictability} helps to remove potential surprises introduced by the machine's decisions, provides insights into uncertainties \cite{DBLP:conf/icse/BencomoB14} and ways in which reliability of the autonomous partner changes over time, under what circumstances it changes, and how its decisions are made. 
Given the importance of the transparency goal, and the many facets of observability and predictability, systems will likely have several associated runtime models. These models tend to use runtime data about the machine's progress and its environment collected using runtime monitors. Based on this, visualization and interactivity support is provided for the analysis phase. \DR primarily supports transparency through the use of several map-based views that depict the current location and status of each UAV, as well as task-based views such as the one presented in the following example.\vspace{4pt}

\noindent{-~\it~Task-Centric Model:~}
In a multi-agent system, the human needs to understand exactly what each individual agent is currently doing and its progress towards the overall mission plan.  The \DR  task-centric view generates a global state transition model combining the states and transitions from all active UAVs' onboard state machines, and then tracks and displays the current task performed by each UAV in the merged state transition model. Progress is then visualized in a UI showing the current task of each UAV as a colored token assigned to a specific node.  This is illustrated in \citefig{state_diagram}, which shows that the Green (G) delivery UAV is {\it on standby}, Red (R) and Orange (O) UAVs are {\it searching}, the Purple (P) UAV is performing {\it surveillance}, and the Blue (B) UAV has {\it detected a victim}.  

\begin{figure}[]
     \centering
     \includegraphics[width=.95\columnwidth]{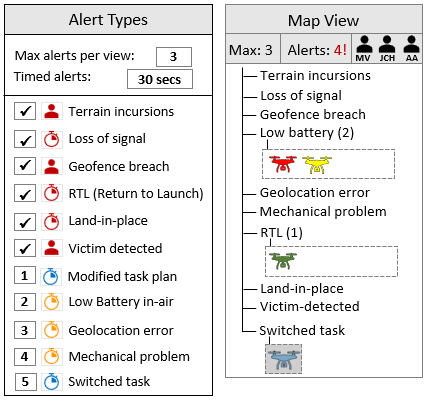}
     \caption{The `Alert Triage' model stores user-created and/or machine-learned prioritization rules, and triages the display of alerts for each view (e.g., Map view on the right).}
        \label{fig:alertsmodel}
        \vspace{-12pt}
\end{figure}

\subsection{Runtime Models that Augment Cognition}

The cognition goal extends far beyond basic transparency and is intended to help humans  understand emergent problems and their causes, provide insights into the decisions and actions taken by the autonomous partner, and allow the user to explore different perspectives and solutions~\cite{liexplanations}. \mkhmt runtime models support cognition goals through enabling the right information to be presented or available to the human at the right time, without overloading their cognitive abilities ~\cite{wortham2017robot, ankit2021, klingberg2009overflowing}.  The two examples we present here focus on \emph{triaging runtime alerts} and \emph{explaining autonomous actions} of the machine by generating human-readable explanations of autonomous behavior; however, our previously presented task-centric model (cf.~\citefig{state_diagram}) provides an additional example, as its interactive GUI also provides interactive and more detailed views of individual UAV's tasks and progress.

\noindent{\it -~Alert Prioritization Model:} The \DR alert prioritization model is designed to avoid the situational awareness ``design demon'' of information overload \cite{endsley2012}, and is built upon a formal meta-model for human-UAV collaborations~\cite{ankit-modre}. It focuses particularly on the HMT goal of \emph{augmenting cognition} with an emphasis on directing human attention to important messages. In \citefig{alertsmodel} alert rules and priorities are shown on the left. They are initially provided as default values by human stakeholders but can be dynamically adapted at runtime, by both the human and the machine. They specify essential alerts which must always be displayed ($\checkmark$), and prioritized alerts (1-5) which will only be displayed if they don't cause the maximum threshold to be exceeded. The \emph{triage} part of the model (e.g., Map View), is dynamically maintained by the system at runtime and is responsible for managing alerts in each active GUI view. It is notified whenever an alert is generated by a runtime model hosted on the UAV or on a GCS microservice. It is also notified by the GUI server whenever a new GUI is activated or deactivated. The alert prioritization model thus builds upon services already available in the basic MAPE-K loop, by collecting, aggregating, and processing the data they produce, in order to support the HMT-focused capability of triaging alerts. 
\vspace{4pt}

\begin{figure}[t]
\begin{subfigure}{1\columnwidth}
\centering
\addtolength{\tabcolsep}{-3pt}
    \begin{tabular}{|c|c|p{6.6cm}|}

	\hline
	\textbf{Type} & \textbf{H/M} & \textbf{Explanation Template} \\
	\hline
	Ext&M & UAV-\{id/color\} identified \{Event\} in the environment. Therefore, adapting \{Action - internal changes\} to \{Rationale\} \\
	\hline
	Ext&H & UAV-\{id/color\} identified \{Event\} in the environment. Therefore, need \{Desired Changes\} to \{Rationale\} \\
	\hline
	Int&M & UAV-\{id/color\} observed \{Event\}. Therefore, \{Action - internal changes\} to \{Rationale\} \\
	\hline
	Int&H & UAV-\{id/color\} observed \{Event\} due to \{cause\}. Therefore, need \{Desired Changes\} to \{Rationale\} \\
	\hline
\end{tabular}
    \caption{Explanation templates for internally and externally triggered adaptation events initiated by either the human (H) or machine (M).}
    \label{tab:exp_template}
    \vspace{8pt}
\end{subfigure}

\begin{subfigure}{1\columnwidth}\centering
    \centering
    \includegraphics[width=1\columnwidth]{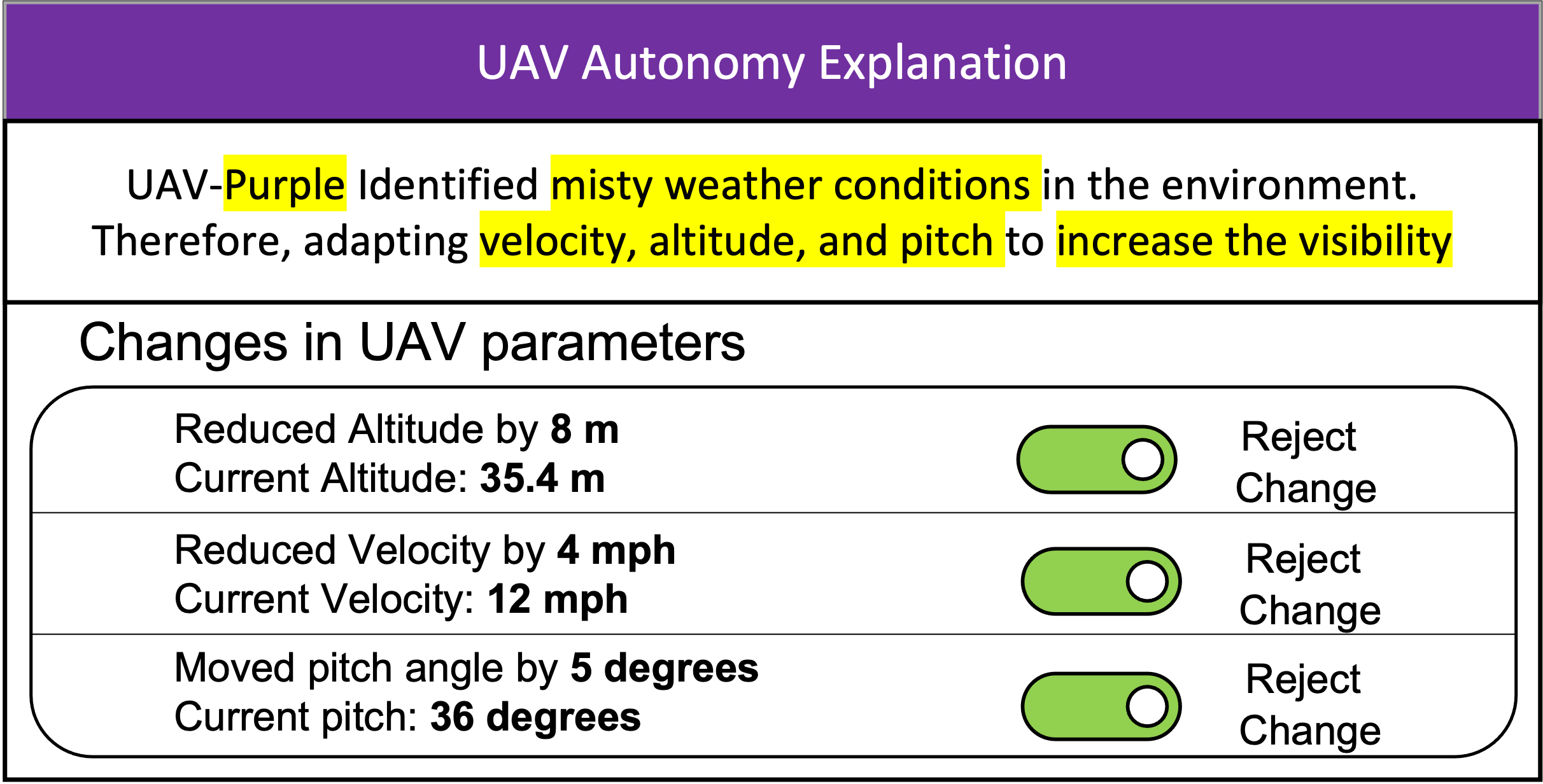}%
\caption{An example explanation generated by the runtime model.}
\label{fig:explanation}
\end{subfigure}
\caption{The Autonomy Explanation Model generates an explanation for all major adaptation decisions.}
\end{figure}

\noindent{\it -~Adaptation Explanation Model:~}
Finally, the explanation model generates explanations for UAV autonomous decisions so that humans can gain insights into the machine's reasoning and assess the appropriateness of individual adaptations~\cite{ankit2021,koo2015did}. These insights potentially strengthen the human's trust and confidence in its machine partner. The predefined \emph{explanation templates} shown in \citetable{exp_template} are used to dynamically generate human-readable textual explanations for all adaptations performed by a UAV. Whenever a UAV self-adapts, it collects three types of information. These are \emph{explanation snippets} describing relevant external events (e.g., ``misty weather conditions'' or ``victim detected''), the UAV's response to the events (e.g., ``reduced altitude by 8 m'' or ``switched to tracking mode''), and finally, the rationale behind those actions (e.g., ``limited visibility'', or ``high confidence in victim sighting''). An example of a weather-related adaptation explanation is shown in Figure \ref{fig:explanation}. Upon receipt of the adaptation message, the runtime model selects the appropriate template and generates the explanation by filling in the missing parts with the data provided by the UAV  \cite{ankit2021}. This model, therefore, utilizes the outputs of existing runtime adaptation models to provide explanations that are critical for HMT.

\subsection{Runtime Models for Coordination} 
\label{sec:coordination}
Coordinating the actions of both humans and machines represents a challenging problem \cite{hey2021} that is exacerbated by the differing cadences of human and machine response times. Successful coordination depends on many factors, including a shared conceptual model of the operating environment, respect for each partner's capabilities, and well-calibrated bidirectional trust. Humans trust the machine to operate autonomously when capable and to request help when needed. Conversely, machines accept interventions from humans and expect feedback when requested, if the human is available. In this context, we present three different runtime models covering the cases of machine-initiated and human-initiated coordination, as well as the particularly challenging case that occurs when humans and machines generate conflicting plans of action. 
\vspace{4pt}

\noindent{\it -~Machine Initiated Coordination:~}
In \DR much of the autonomous behavior of the UAV is supported by its onboard CV capabilities (cf.~\citefig{cv})~\cite{DBLP:journals/corr/abs-2103-15053}. For example, when searching for a drowning victim, the UAV uses CV to continuously analyze the video stream and detect objects classified as ``person''. For each detected object, the CV module generates two scores. The \emph{confidence} score represents the probability that the object is correctly classified, while the \emph{reliability} score accounts for any {\it uncertainty} arising from image noise or mismatch between the current context and the training data ~\cite{pouget2016confidence}. Based on learned threshold values, the UAV uses these scores to decide whether it can autonomously decide on its actions (e.g., track the object vs. continue to search) or should request help from the human. Both the CV model and the related human-machine coordination mechanism are supported by runtime models which take a coordination specification as input (cf.~Algorithm 1), instantiate a simple state machine (managed by a microservice on the GCS), and use the existing messaging system to choreograph human and machine tasks. In closely related work Li~\etal, also proposed a form of choreography that provides users with advanced knowledge of tasks they would be requested to participate in \cite{hey2021}.\\

\begin{figure}[t]
  \centering
  \includegraphics[width=1\columnwidth]{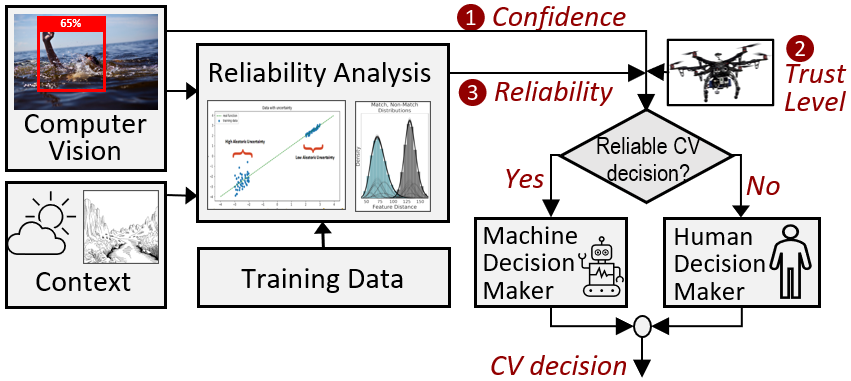} 
  \caption{The UAV's Computer Vision component considers confidence,  reliability, and calibrated trust levels, to determine when to request help from the human partner \cite{DBLP:journals/corr/abs-2103-15053}.} 
   \label{fig:cv}
\end{figure}

\begin{algorithm}
\SetAlgoLined
  \If{Object detected at low reliability}{
    UAV raises alert and requests help from human\;
    \uIf{Human is available and responsive}{
        Human evaluates video stream, makes decision, and selects \texttt{CONFIRM} or \texttt{REJECT} option\;
        \uIf{Human confirms victim sighting}{
           UI Server sends \texttt{CONFIRMATION} message to UAV;}
        \Else{Human refutes sighting\;
           UI Server sends \texttt{REFUTATION} message to UAV;}
     }   
    \If{no response from human within waiting\_period}{
        \texttt{NO RESPONSE} message sent to UAV\; 
        `human failure to respond' event is logged\;
        Responsibility reverts to UAV\;}
   }
 \caption{Human-Machine coordinated decision-making addressing Computer Vision reliability problems }
 \label{pattern:3}
\end{algorithm}
\vspace{-10pt}

\noindent{\it -~Human Initiated Coordination Models:~}
In order for the human to engage in meaningful interactions with the machine, they must have high degrees of situational awareness and a clear understanding of what they can and cannot do at any point during a mission. Furthermore, the different cadences of human and machine response times within \mkhmt create a significant coordination challenge resulting in problems such as the human making plans based on stale data or intervening after the machine has already completed an action.  An example of a relevant runtime model determines currently available human interaction options and dynamically adapts each active GUI to activate and deactivate widgets (e.g., icons, buttons, menu options) according to the ways that humans can feasibly interact with the machine given the current state of the mission. For example, if the UAV is currently in active search mode, it is reasonable for the human to request a view of the UAV's annotated video stream; however, this option should not be available if the UAV is in RTL mode with cameras turned off to preserve limited power. A suitable affordance (e.g., a button) should be activated and deactivated accordingly). Tracking these currently available human actions is handled by a dedicated runtime model.
\vspace{4pt}

\noindent{\it-~Mitigating Tug-of-War Scenarios:~} Finally, a significant challenge in HMT is reconciling potentially conflicting actions taken by the human and machine, a problem that is again exacerbated by the different operating speeds of the machine and human. As the machine has the advantage of faster cadence, it will often win any disagreements with potentially devastating results, as illustrated by the case of the Lion Air Crash (cf.,~\citesec{planning}). Similar scenarios play out across other domains. For example, a UAV may place itself close to the river to collect better imagery of the riverbank. On a sunny day,  reflections from the light on the water could impact the UAV's sensors causing sudden altitude fluctuations and triggering a land-in-place failsafe mechanism to activate. An alerted human might quickly intervene to prevent the UAV from landing in the river directing it to ascend to a safe altitude; however, once the UAV's autonomy kicks back into gear, the whole cycle could repeat itself -- causing a tug-of-war with respect to the ideal altitude placement. HMT systems must allow humans and/or machines to detect and break interwoven cycles of human and machine actions that are indicative of a tug-of-war. As a simple solution, detected cycles could be broken by curtailing the UAV's autonomy until reestablished by the human.

\subsection{Integrating the RunTime Data Probes}
While we have presented these six example models as independent entities, many of them share common data inputs (e.g., data read from sensors and software probes), or rely on data/outputs from another runtime model. For example, in order to generate meaningful explanations where multiple UAVs are involved, information is collected from the individual UAV onboard runtime models. 
The Autonomy Explanation Model relies upon ``adaptation notification events'' provided by each individual UAV, which in turn employs their own state machine. This information is then aggregated and post-processed, so that the Alert Prioritization Model (cf.~\citefig{alertsmodel}) can select and display critical alerts from individual UAVs or global alerts from the system. 
While both the knowledge base, and (parts) of the runtime models are distributed across the respective agents they share data as needed. This means that the UAV retains full autonomy when a person is detected, and adaptation decisions are made (onboard), but at the same time raises alert events which support decision-making of the human partner.
This dependence on accurate and appropriate information between the different models calls for thorough design and planning to ensure that the right information is available at the right time in a potentially resource-constrained operating environment.

\section{Analysis and Take-Aways}
\label{sec:analysis}
In this paper, we have proposed augmenting the MAPE-K loop with support for HMT. \mkhmt assumes that both humans and machines are capable of autonomous behavior and decision-making, and that mission goals are achieved jointly through an interactive partnership. Based on our own experiences in applying \mkhmt to the \DR system, we lay out an initial process for designing and deploying \mkhmt.

\begin{enumerate}[leftmargin=*]
  \setlength\itemsep{.3em}
    \item \emph{Stakeholder Identification:~} HMT systems inherently involve humans, and therefore, it is important to identify CRACK (Collaborative, Representative, Authorized, Committed, Knowledgeable) \cite{DBLP:conf/icse/BoehmT04} stakeholders serving as direct users and domain experts.  Engaging with stakeholders, who will become the human partners, helps to uncover interactions and expectations users have on the system, for example, by carefully exploring the human-machine interactions related to mission-related scenarios  ~\cite{sutcliffe1998supporting}.  
    
    \item \emph{Elicitation of HMT requirements:~} Once stakeholders are identified, requirements must be elicited. As a starting point,  McDermott~\etal~\cite{HMT2} has described a detailed elicitation process, supported by a list of key questions associated with each of the HMT factors \cite{HMT3}. For example, to understand ``predictability'' requirements, analysts must discover (a) automation goals, abilities, and limitations, (b) how the human partner's goals and priorities are tracked, (c) reliability of different automated tasks within different contexts, and (d) the types of changes that are expected to occur and trigger subsequent adaptations.   
    
    \item \emph{Requirements Analysis and Specification:~} The elicited requirements are subsequently analyzed to negotiate and reconcile trade-offs, and to identify and specify requirements that support human-machine interactions~\cite{ruhe2003trade}. While our process is agnostic to specific techniques, the domains in which MAPE-K operates generally dictate that the requirements process includes a rigorous safety analysis (e.g.,  \cite{leveson2011, denney2012heterogeneous, vierhauser2021hazard}), and determines that requirements should be specified sufficiently formally to capture timing and other performance constraints and/or to establish formal goal models. In many cases, the requirements specification (e.g., Goal Models or state transition diagrams) provides the foundation for the respective runtime models~\cite{nelly19,anda2019arithmetic,brings2020goal}.
    
    
    \item \emph{Design and Integration of HMT runtime models:~} \mkhmt emphasizes the importance of HMT-related runtime models. We, therefore, start the design process by taking an inventory of existing runtime models, identifying gaps where HMT requirements are not adequately supported by existing models, designing new runtime models as needed, and then finally composing models into workflows to service each HMT requirement. This involves assessing \emph{who} (machine or human-role), \emph{when},  and \emph{where} each model will be used and updated, and \emph{what} data sources are required as inputs (e.g., probes, message subscriptions, or human-initiated data and events). Furthermore, as actions are performed at different speeds, the required refresh frequencies must be determined for each constituent element of each runtime model, and a system-wide plan established so that each collected data attribute satisfies refresh frequencies of all relevant models.  
    
    \item \emph{User Interface Design:~} User interfaces need to be designed to provide interactive support for humans.  Depending on the data that is displayed, or the input that is expected from the human, this may include GUIs, or hardware interfaces such as Joysticks or radio controllers. Mission- and safety-critical information needs to be provided to the human in a timely manner without creating cognitive overload, and must not be ``hidden'' in sub-menus or views that require multiple steps to access~\cite{endsley2012}.
 
    \item \emph{Implementation, Testing, and Deployment:~} The final steps involve implementing the system, verifying that required runtime models, UIs, and supporting features are implemented as intended, and finally validating that the deployed system satisfies its stated requirements and supports the desired HMT. Further discussion of these steps is outside the scope of this paper.
\end{enumerate}

\section{Threats to Validity}
\label{sec:threats}
Our work is subject to three primary threats to validity. 
First, all of the runtime models described in this paper are designed to support HMT in our own \DR system, representing a multi-agent, multi-human system operating in a mid-level safety-critical domain \cite{DBLP:journals/tse/VierhauserBWXCH21,uavsafety}. As  types of  interactions are influenced by the operating domain, the human-machine partnerships in \DR may differ significantly from other application domains -- for example, those with a single operator or a single machine, or operating in a highly safety-critical domain. On the other hand, instead of proposing a specific set of HMT-related runtime models, \mkhmt provides a simple process for identifying, developing, and integrating context-specific models in support of  transparency, augmented cognition, and coordination goals, which are known to be applicable across diverse operating environments \cite{HMT,HMT2,HMT3}. Further, our process could be easily extended -- for example, by integrating a more formal safety analysis for more critical domains. Our future work will apply this process to more varied systems.

Second, while our runtime models have been prototyped in various forms, only some have been integrated into the full version of \DR. As such, this paper represents a vision, supported by concrete examples. Future work will focus on concrete implementation details. For example, while we have discussed the creation of shared data probes that service the refresh frequencies of all runtime models, and have developed monitoring capabilities to achieve this, this aspect of the work needs further investigation. 

Finally, while \DR has been deployed in the physical world, formal evaluations of HMT user interfaces have relied upon user studies  conducted in \DRx's simulator (e.g., \cite{Agrawal20,ankit2021,vierhauser2018monitoring}). However, the \DR GUI is identical for both physical and simulated UAVs, and our past experiences have shown that findings from these user studies have been effective when deployed in physical field tests.  Nevertheless, further experiments targeted specifically at HMT need to be conducted in real-world settings where humans collaborate with UAVs under far more noisy, volatile, and potentially stressful conditions.

\section{Related Work}
\label{sec:related}
A significant body of work, primarily in the HCI  community has focused on designing UIs to support situational awareness of Cyber-Physical Systems (e.g.,~\cite{demonhunt,Agrawal20, frobt18,OutOfLoop_biondi201880,regis14, ankit-modre}). The focus has primarily been on enabling users to perceive, understand, and make effective decisions~\cite{endsley2012,endsley2017autonomous}. While this related work supports  HMT goals, it puts little emphasis on integration with underlying runtime models which are needed in order to deliver accurate, timely, and often aggregated data for use in the UIs.  Kephart advocated for increased interactivity between humans and users in adaptive decision-making \cite{keynote-Kephart}; however, their examples were all taken from domains in which humans had plenty of time to consider their decisions. Integrating HMT into the MAPE-K loop for real-time robotics systems introduces additional, and very  challenging, timing constraints.


In the HMT domain, researchers have explored many facets of human-machine teaming. For example, Klein~\etal~\cite{klien2004ten} identified challenges associated with achieving shared goals, preventing breakdowns in team coordination, and fostering  communication and collaboration. Furthermore, Schmid~\etal~\cite{schmid20} studied ways to adjust system automation in complex, safety-critical environments in order to better support human operators. While their work has significant relevance to \mkhmt, it perceives humans as ``operators'' rather than true partners.

Calhoun~\etal~\cite{Calhoun2017OperatorAutonomyTI} proposed a flexible architecture that allows the degree of automation to vary according to the human's current engagement and workload. These goals are reflected in our discussion on adaptation in \mkhmt, which embraces the notion of adapting for improved human collaboration and performance. To increase context awareness in order to better engage humans in the decision-making process, Li~\etal~\cite{hey2021} proposed a formal framework based on probabilistic reasoning to determine when advanced notifications are useful for humans interacting with self-adaptive systems. Their work specifically addresses the cadence problem in which humans may be required to respond quickly but require ``thinking time''. However, their evaluation was conducted in a robotics goods delivery domain, which has a lower decision cadence than a multi-UAV system. 

Finally, several papers have explored self-adaptation and/or human interactions in the UAV domain.  Vierhauser~\etal~\cite{vierhauser2021hazard} identified human-UAV interaction hazards and potential mitigations, whilst Miller~\etal~\cite{miller2019} explored different techniques by which users could interact with multiple UAVs.  However, these papers focus on  Human-on-the-loop rather than HMT. Other researchers have proposed UAV-related self-adaptation frameworks. For example, Braberman~\etal~\cite{braberman2015morph,extended_MORPH} presented a MAPE-K reference architecture for unmanned aerial vehicles, while Yu~\etal~\cite{livebox} proposed a self-adaptive framework for UAV forensics. Neither of these explored the HMT aspects of adaptation. Finally, in more closely related work Lim~\etal~\cite{lim21} explored ways to adapt human-robot interactions in a multi-UAV system, with a focus on modulating automation support according to the cognitive states of the human operator. This creates a subtle, but important difference from our \mkhmt goal, as it centers primarily around the needs of the operator rather than optimizing teamwork goals.

\section{Conclusion}
\label{sec:conclusion}
This paper has described the techniques and process we have used to augment MAPE-K in order to address the three HMT goals of transparency, augmented cognition, and coordination. We have described how MAPE-K provides a meaningful self-adaptation framework in which humans and machines collaborate together, and have mapped HMT factors to the MAPE-K phases and then used a series of examples to illustrate how carefully designed runtime models enable meaningful support for human-machine partnerships. Rather than diminishing the autonomy of machines, HMT draws upon the autonomous abilities of both humans and machines to deliver an even stronger solution realized through developing meaningful teamwork.

In conducting this work we have identified two key challenges. The first stems from the very different cadences at which humans and machines operate. This creates the potential for humans to make decisions based upon stale data, and for directives interjected into the machine's plans to inadequately reflect the current state of the system. Existing HMT solutions, such as turn-taking~\cite{chao2016timed}, are only effective in scenarios which can tolerate slower response times. In \DR we partially address this challenge by dynamically adapting the UIs to only reflect  currently available human interaction options, and by implementing a reliable messaging system which ensures that messages are only sent to the UAV when it is in a state to handle them; however, the problem is a complex one which warrants further exploration.
The second challenge relates to designing and supporting the integrated \mkhmt environment. As different runtime models are required for effective human-machine teaming, and these need to be identified, designed, integrated, deployed, and effectively maintained at runtime. 

Even beyond these challenges, much additional work is needed in order to realize the vision of trusted, coordinated, and accountable teams of humans and machines. Our ongoing work will therefore investigate solutions for addressing these challenges, create a fully integrated \mkhmt environment for evaluation purposes, conduct field-tests with physical UAVs, and explore the deployment of \mkhmt across a much broader set of different domains.

\section*{Acknowledgment}
The work in this paper has been funded by the US National Science Foundation (NSF) under Grant\# CNS-1931962 and also by the  Linz Institute of Technology (LIT-2019-7-INC-316).
\bibliographystyle{ACM-Reference-Format}
\balance
\interlinepenalty=10000
\bibliography{SEAMS.bib} 
\end{document}